\documentclass[runningheads]{llncs}
\usepackage{graphicx}
\usepackage{balance}
\usepackage{amssymb}
\usepackage{epstopdf}
\usepackage{listings}
\usepackage{amsmath}
\usepackage{bigstrut}
\usepackage{tabularx}
\usepackage{booktabs}
\usepackage[font=small,skip=0pt]{caption}

\usepackage{tikz}
\usepackage{pgfplots}
\usepackage{filecontents}
\pgfplotsset{width=8cm,compat=1.9}
\usetikzlibrary{pgfplots.dateplot}
\usepackage{pgfplotstable}

\usepackage[ruled,vlined,linesnumbered,lined,commentsnumbered]{algorithm2e}
\usepackage{fancyhdr}
\fancypagestyle{specialfooter}{%
  \fancyhf{}
  
  \fancyfoot[L]{\hspace{2mm}Copyright \textcopyright2019 for this paper by its authors. Use permitted under Creative Commons License Attribution 4.0 International (CC BY 4.0). FIRE 2019, 12-15 December 2019, Kolkata, India.}
}

\begin{document}
\title{QutNocturnal@HASOC'19: CNN for Hate Speech and Offensive Content Identification in Hindi Language}
%
%
\author{Md Abul Bashar\orcidID{0000-0003-1004-4085} \and
Richi Nayak\orcidID{0000-0002-9954-0159}}
\authorrunning{Bashar \& Nayak}
%
\institute{School of Electrical Engineering and Computer Science \\
Queensland University of Technology, Brisbane, Australia\\
\email{\{m1.bashar, r.nayak\}@qut.edu.au}}
\maketitle              
\begin{abstract}
We describe our top-team solution to Task 1 for Hindi in the HASOC contest organised by FIRE 2019. The task is to identify hate speech and offensive language in Hindi. More specifically, it is a binary classification problem where a system is required to classify tweets into two classes: (a) \emph{Hate and Offensive (HOF)} and (b) \emph{Not Hate or Offensive (NOT)}. In contrast to the popular idea of pretraining word vectors (a.k.a. word embedding) with a large corpus from a general domain such as Wikipedia, we used a relatively small collection of relevant tweets (i.e. random and sarcasm tweets in Hindi and Hinglish) for pretraining. We trained a Convolutional Neural Network (CNN) on top of the pretrained word vectors. This approach allowed us to be ranked first for this task out of all teams. Our approach could easily be adapted to other applications where the goal is to predict class of a text when the provided context is limited.
\end{abstract}

\thispagestyle{specialfooter}
\section{Introduction}
The ``Hate Speech and Offensive Content Identification in Indo-European Languages'' track\footnote{https://hasoc2019.github.io/call\_for\_participation.html} (HASOC) is one of the tracks in FIRE 2019
conference\footnote{http://fire.irsi.res.in/fire/2019/home} \cite{hasoc2019overview}. Task 1 in this track is identification of hate speech and Offensive (HOF) language in English, German and Hindi in social media posts. In this paper, we describe our approach to the solution of Task 1 in Hindi. The goal is to label a tweet written in Hindi as HOF if it contains any form of non-acceptable language such as hate speech, aggression or profanity; otherwise it is labelled as NOT. There has been significant research on hate speech and offensive content identification in several languages, especially in English \cite{bashar2018misogynistic,badjatiya2017deep,davidson2017automated,zampieri2019semeval,zampieri2019predicting}. However, there is a lack of work in most other languages.  
People are now realising the urgency of such research in other languages.
Recently, SemEval 2019 Task 5 \cite{basile2019semeval} was carried out on detecting hate speech against immigrants and women in Spanish and English messages extracted from Twitter, GermEval  Share Task \cite{wiegand2018overview} was carried out on the Identification of Offensive Language in German language tweets, and TRAC-1 \cite{kumar2018benchmarking} conducted a shared task on aggression identification in Hindi and English. Therefore, HASOC Task 1 for Hindi intends to find out the quality of hate speech and offensive content identification technology in Hindi.

The training dataset is comprised of 4665 labelled tweets in Hindi. The training dataset is created from Twitter and participants are allowed to use external datasets for this task. In the competition setup, the testing dataset is comprised of 1319 unlabelled tweets that were also created from Twitter. The testing dataset and leaderboard were kept unknown to participants until the results were announced. Competitors had to split the training set to get validation set and use the validation set through the competition to compare models. The testing set was only used at the end of the competition for the final leaderboard. 

Th proposed approach relies on very little feature-engineering and preprocessing as compared to many existing approaches. Section \ref{sec:winning_approach} discusses our top-ranked model building approach. It consists of two steps: (a) pretraining word vectors using a relevant collection of unlabelled tweets and (b) training a Convolutional Neural Network (CNN) model using the labelled training set on top of the pretrained word vectors. Section \ref{sec:alternative_models} describes other sophisticated alternative models that we tried. Though these models did not perform as good as compared to our winning model in this track, their performance provides further insight into how to use machine learning models for identifying hate speech and offensive language in Hindi. Section \ref{sec:experimental_results} provides experimental results comparing and analysing our various models both on testing set and validation set. The source code of our model can be found online at \cite{CodeQutNocturnalHasoc2019}. 

\section{The Winning Model: QutNocturnal}
\label{sec:winning_approach}

\subsection{Data Collection}

\subsubsection{Labelled Contest Dataset}
The goal of Task 1 for Hindi is to predict the class (HOF or NOT) of a given tweet written in Hindi. Out of 4665 labelled tweets in the training set, 2469 (52.92\%) are HOF and 2196 (47.07\%) are NOT. We randomly kept 20\% of training data for validation set. We used ten cross validation in the remaining training set for hyper parameter setting.

\subsubsection{Unlabelled External Dataset}
\label{sec:external_data}
It is a difficult task to separate abusive tweets from tweets that are sarcastic, joking, or contained abusive keywords in a non-abusive context \cite{bashar2018misogynistic}. Lexical detection methods tend to have low accuracy \cite{davidson2017automated,xiang2012detecting} because they classify a tweet as abusive if it contains any abusive keywords. Also tweets are significantly noisy and do not follow a standard language format. For example, words in tweets are often misspelled, altered, written in Roman letters, include local dialects or foreign languages. To transfer the knowledge of these contexts to the CNN based deep learning model, we pretrain word vectors using 0.5 million relevant tweets. More specifically, we collected 4,94,311 random tweets in Hindi (i.e. topic of discussion can be anything) using TrISMA\footnote{https://research.qut.edu.au/dmrc/projects/trisma-tracking-infrastructure-for-social-media-analysis/} and 5251 sarcasm tweets in Hinglish \cite{mathur2018detecting} (i.e. sarcasm in Hindi language but written in Roman letters) from \cite{swami2018corpus} for pretraining.

\subsubsection{Preprocessing}
We de-identified person occurrence (e.g. @someone) with \emph{xxatp}, url occurence with \emph{xxurl}, source of modified retweet with \emph{xxrtm} and source of unmodified retweet with \emph{xxrtu}. We fixed the repeating characters (e.g. goooood) in word and removed common invalid characters (e.g. $<br/>$, $<unk>$, $@-@$, etc). We used html unescape to replace hexadecimal escape sequences with the character that it represents. We used multi-language spaCy module\footnote{https://spacy.io/models/xx} to lemmatize words and a lightweight stemmer for Hindi language \cite{ramanathan2003lightweight} for stemming the words. 

\subsection{Word Embedding}
\begin{figure}[htb]
	\centering
	\scriptsize
	\includegraphics[width=0.8\textwidth]{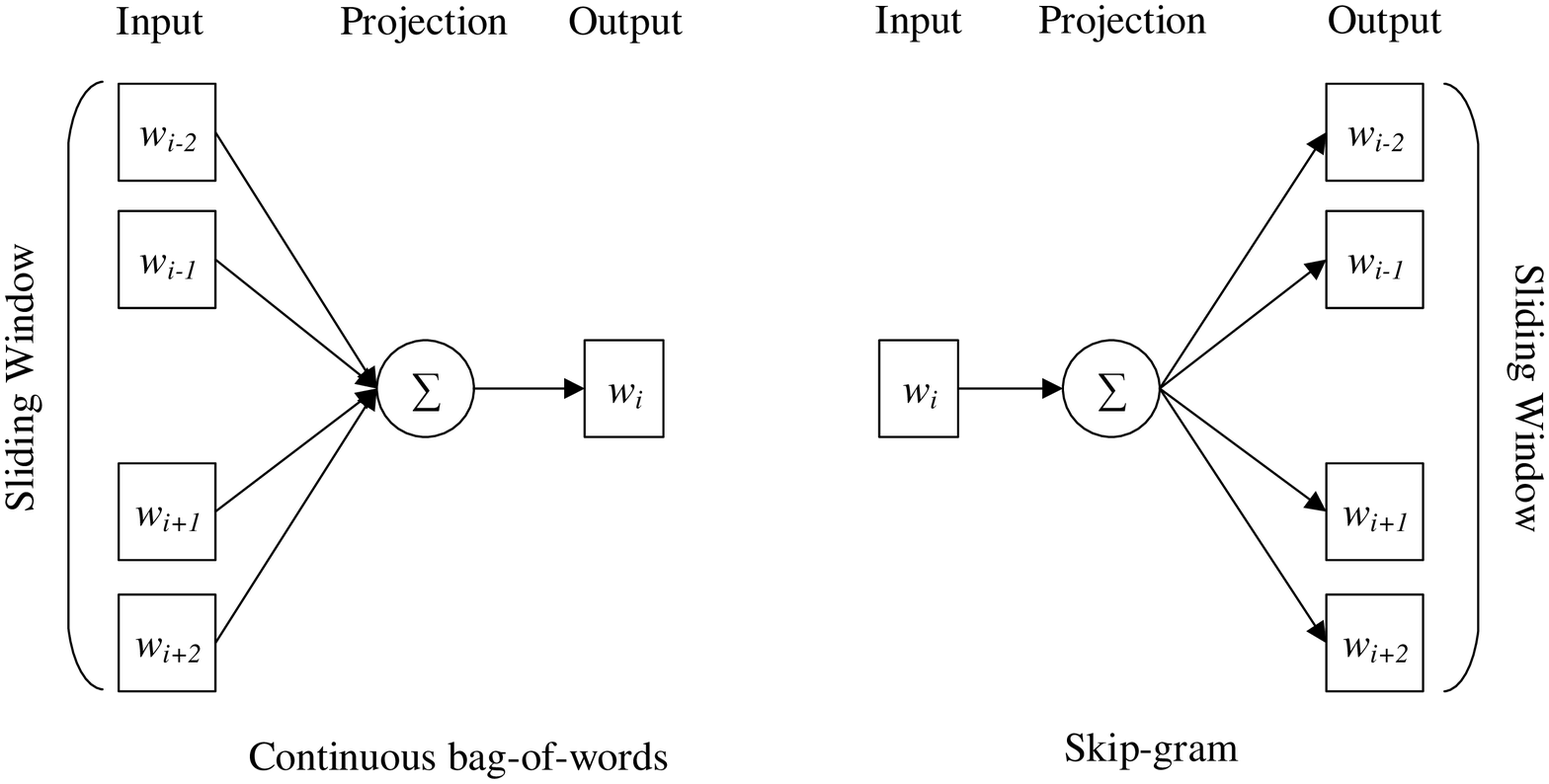}
	\caption{Continuous Bag-of-Words and Skip-gram Word Embedding Models \cite{bashar2018misogynistic}}
	\label{fig:Word2vec_Architectures}
\end{figure}

Embedding models quantify semantic similarities between words based on their distributional property that a word is characterised by the company it keeps.
These models quantify semantic properties of words by mapping co-occurring words close to each other in an Euclidean space. Given a sizeable corpus, these models can effectively learn a high-quality word embedding from the co-occurrence of words in the corpus. Word embedding maps each word from the vocabulary to a vector of real numbers. Mikolov et al. \cite{mikolov2013distributed} proposed two popular models for word embedding based on the feed-forward neural network: Skip-gram and Continuous Bag-of-Words as shown in Figure \ref{fig:Word2vec_Architectures}.

In embedding models, a sliding window of a fixed size moves along the text of a corpus. For a given position of the sliding window, let the word in the middle is current word $w_i$ and the words on its left and right within the sliding window are context words $C$. The continuous bag-of-words model predicts the current word $w_i$ from the surrounding context words $C$, i.e. $p(w_i|C)$. In contrast, the skip-gram model uses the current word $w_i$ to predict the surrounding context words $C$, i.e. $p(C|w_i)$. In Figure \ref{fig:Word2vec_Architectures}, for example in this corpus, if the current position of a running sliding window contains the phrase \emph{tum sirf chutiya kat ti ho}. In continuous bag-of-words, the context words \{tum, sirf, kat, ti, ho\} can be used to predict the current word \{chutiya\}, whereas, in skip-gram, the current word \{chutiya\} can be used to predict the context words \{tum, sirf, kat, ti, ho\}. 

The objective of model training is to find a word embedding that maximises $p(w_i|C)$ or $p(C|w_i)$ over a corpus. In each step of training, each word is either (a) pulled closer to the words that co-occur with it or (b) pushed away from all the words that do not co-occur with it. A\textit{ softmax} or \textit{approximate softmax} function can be used to achieve this objective \cite{mikolov2013distributed}. At the end of the training, the embedding brings closer not only the words that are explicitly co-occurring in a training dataset, but also the words that implicitly co-occur. For example, if $w_1$ explicitly co-occurs with $w_2$ and $w_2$ explicitly co-occurs with $w_3$, then the model can bring closer not only $w_1$ to $w_2$, but also $w_1$ to $w_3$. 

We use the continuous bag-of-words model in this contest as this model is faster and has a slightly better accuracy for the words that appear frequently based on our experimental results. 
We implemented this model using the module Word2Vec in Gensim Python library. We set the word vector dimension to 200, minimum word count to 2, number of iteration in pretraining to 10, sliding window size to 5 and maximum vocabulary count to 0. 
We run this model on the unlabelled external dataset described in Section \ref{sec:external_data} to get the pretrain word vectors. Our pretrained word vectors and corresponding python code to use them in classifier are available online at \cite{CodeQutNocturnalHasoc2019}.

\subsection{Model Architecture}
\begin{figure}[htb]
	\centering
	\scriptsize
	\includegraphics[width=0.8\textwidth]{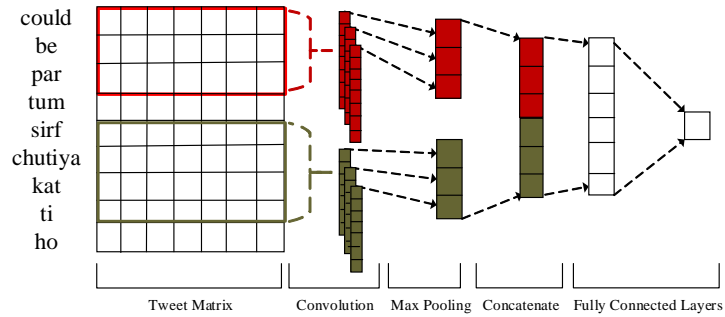}
	\caption{Architecture of our top-ranked CNN Model for the Hate Speech and Offensive Content Identification track in Hindi Language \cite{bashar2018misogynistic}}
	\label{fig:ModelArchitecture}
\end{figure}

The proposed architecture of our top-ranked model CNN to identify hate speech and offensive language in Hindi is given in Figure \ref{fig:ModelArchitecture}. This is an empirically customised and regulated version of the architecture that we have used in our prior work of misogynistic tweets identification on Tweeter \cite{bashar2018misogynistic}. In this architecture, we use word embedding to represent each word $w$ in an $n$-dimensional word vector $\mathbf{w} \in \mathbb{R}^n$. We represent a tweet $t$ with $m$ words as a matrix $\mathbf{t} \in \mathbb{R}^{m \times n}$. We apply convolution operation to the tweet matrix with one stride. Each convolution operation applies a filter $\mathbf{f}_i \in \mathbb{R}^{h \times n}$ of size $h$. Empirically, based on the accuracy improvement in ten-fold cross validation, 256 filters are used for $h \in \{3,4\}$ and 512 filters for $h \in \{5\}$. The convolution is a function $\mathbf{c}(\mathbf{f}_i, \mathbf{t}) = r(\mathbf{f}_i \cdot \mathbf{t}_{k:k+h-1})$, where $\mathbf{t}_{k:k+h-1}$ is the $k$th vertical slice of the tweet matrix from position $k$ to $k+h-1$, $\mathbf{f}_i$ is the given filter and $r$ is a Rectified Linear Unit (ReLU) function \cite{nair2010rectified}. The function $\mathbf{c}(\mathbf{f}_i, \mathbf{t})$ produces a feature $c_k$ similar to $n$Grams for each slice $k$, resulting in $m-h+1$ features. The max-pooling operation \cite{tolias2015particular} is applied over these features and the maximum value is taken, i.e. $ \hat{c}_i = max(\mathbf{c}(\mathbf{f}_i, \mathbf{t}))$. Max-pooling captures the most important feature for each filter. As there are a total of 1024 filters (256+256+512) in the proposed model, the 1024 most important features are learned from the convolution layer. 

Then, we pass these features to a fully connected hidden layer with 256 perceptrons that use the ReLU activation function. This fully connected hidden layer learns the complex non-linear interactions between the features from the convolution layer and generates 256 higher level new features. Finally, we pass these 256 higher level features to the output layer with single perceptron that uses the sigmoid activation function. The perceptron in output layer generates the probability of the tweet being HOF or NOT. 

In this architecture (Figure \ref{fig:ModelArchitecture}), a proportion of units are randomly dropped-out from each layer except the output. This is done to prevent co-adaptation of units in a layer and to reduce overfitting. We set 50\% units droppedout from the input layer, the filters of size 3 and the fully connected hidden layer based on best empirical results. Only 20\% units are droppedout from the filters of size 4 and 5. Python code for this model is available online at \cite{CodeQutNocturnalHasoc2019}. 

\setlength{\textfloatsep}{0pt}

\section{Alternative Models}
\label{sec:alternative_models}
We have implemented eight other models in addition to the winning CNN model to see the performance of hate speech and offensive language detection in Hindi.
\begin{itemize}
	\item Long Short-Term Memory Network (LSTM) \cite{hochreiter1997long}. We implement LSTM with 100 units, 50\% dropout, binary cross-entropy loss function, Adam optimiser and sigmoid activation. 
	\item Feedforward Deep Neural Network (DNN) \cite{glorot2010understanding}. We implement DNN with five hidden layers, each layer containing eight units, 50\% dropout applied to the input layer and the first two hidden layers, softmax activation and 0.04 learning rate. We manually tuned hyper parameters of all neural network based models (CNN, LSTM, DNN) based on cross-validation. 
	\item Non NN models including Support Vector Machines (SVM) \cite{hearst1998support},  Random Forest (RF) \cite{liaw2002classification}, XGBoost (XGB) \cite{chen2016xgboost},
	Multinomial Naive Bayes (MNB) \cite{lewis1998naive}, k-Nearest Neighbours (kNN) \cite{weinberger2009distance} and Ridge Classifier (RC) \cite{hoerl1970ridge}. We automatically tune hyper parameters of all these models using ten-fold cross-validation and GridSearch from scikit-learn. Among all the models, only CNN and LSTM use transfer learning.
\end{itemize}

\section{Experimental Results}
\vspace{-3mm}
\label{sec:experimental_results}
A total of nine machine learning models, including the winning customised CNN model, were trained to identify hate speech and offensive language in Hindi. We used transfer learning of word vectors for both CNN and LSTM. The word vectors were pre-trained on a collection of relevant tweets and tuned with the training dataset during the model training.

\vspace{-2mm}
\subsection{Results}
\vspace{-2mm}
The experimental results comparing models in custom validation set are given in Table \ref{tab:comp_models_val}. 
\begin{table}[h]
	\centering
	\scriptsize
	\caption{Model Comparison Results in Custom Validation Set}
	\begin{tabular}{lccccccccc}
		\toprule
		& \multicolumn{9}{c}{Macro Average of Classes} \\
		\midrule
		& CNN   & DNN   & kNN   & LSTM  & MNB   & RF    & RC    & SVM   & XGB \\
		precision & \textbf{0.83} & 0.72  & 0.61  & 0.79  & 0.76  & 0.74  & 0.73  & 0.68  & 0.74 \\
		recall & \textbf{0.82} & 0.72  & 0.56  & 0.78  & 0.75  & 0.74  & 0.72  & 0.61  & 0.75 \\
		f1-score & \textbf{0.81} & 0.72  & 0.51  & 0.78  & 0.75  & 0.74  & 0.72  & 0.58  & 0.74 \\
		support & 933   & 933   & 933   & 933   & 933   & 933   & 933   & 933   & 933 \\
		\midrule
		&       &       &       &       &       &       &       &       &  \\
		& \multicolumn{8}{c}{Weighted Average of Classes}                              &  \\
		\midrule
		& CNN   & DNN   & kNN   & LSTM  & MNB   & RF    & RC    & SVM   & XGB \\
		\midrule
		precision & \textbf{0.84} & 0.72  & 0.61  & 0.79  & 0.76  & 0.74  & 0.73  & 0.68  & 0.75 \\
		recall & \textbf{0.82} & 0.72  & 0.58  & 0.78  & 0.76  & 0.74  & 0.73  & 0.63  & 0.74 \\
		f1-score & \textbf{0.81} & 0.72  & 0.52  & 0.78  & 0.75  & 0.74  & 0.73  & 0.58  & 0.75 \\
		support & 933   & 933   & 933   & 933   & 933   & 933   & 933   & 933   & 933 \\
		&       &       &       &       &       &       &       &       &  \\
		& \multicolumn{8}{c}{Accuracy}                                  &  \\
		\midrule
		& CNN   & DNN   & kNN   & LSTM  & MNB   & RF    & RC    & SVM   & XGB \\
		& \textbf{0.82}  & 0.72  & 0.58  & 0.78  & 0.76  & 0.74  & 0.73  & 0.63  & 0.74 \\
		\bottomrule
	\end{tabular}%
	\label{tab:comp_models_val}%
\end{table}%
The detailed results of the winning CNN model in test dataset are given in Table \ref{tab:cnn_test}.\footnote{In the absence of any other information except the email message about the top-team performance, we are not able to provide the comparative results with other submitted team results. We will update this table with the rest of the team performance, once we receive information from the track organisers.} 
\begin{table}[h]
	\centering
	\scriptsize
	\caption{Detailed Results of Winning Model CNN in Test Dataset}
	\begin{tabular}{lcccc}
		\toprule
		 &  & Confusion Matrix &  &  \\
		 \midrule
		\multicolumn{1}{c}{HOF} & \multicolumn{1}{c}{NOT} & \multicolumn{1}{c}{} &       &  \\
		\midrule
		\multicolumn{1}{c}{446} & \multicolumn{1}{c}{159} & \multicolumn{1}{l}{HOF} &       &  \\
		\multicolumn{1}{c}{80} & \multicolumn{1}{c}{633} & \multicolumn{1}{l}{NOT} &       &  \\
		\bottomrule
		&       &       &       &  \\
		&       &  Class Wise Performance     &       &  \\
		\toprule
		& Precision & Recall & F$_1$-score & Support \\
		\midrule
		HOF   & 0.85  & 0.74  & 0.79  & 605 \\
		NOT   & 0.8   & 0.89  & 0.84  & 713 \\
		\midrule
		Accuracy &       &       & 0.82  & 1318 \\
		Macro avg & 0.82  & 0.81  & 0.81  & 1318 \\
		Weighted avg & 0.82  & 0.82  & 0.82  & 1318 \\
		\bottomrule
	\end{tabular}%
	\label{tab:cnn_test}%
\end{table}%

\subsection{Analysis of the results}
Experimental results in both validation and test set show that CNN outperforms all other models. 
CNN is able to outperform LSTM and other baseline models because of the specific nature of tweets. For example, tweets can be super condensed and indirect texts (e.g. satire), may not follow the standard sequence of the language and be full of noise. 

Traditional models (e.g. SVM, XGBoost, RF, kNN, etc.) are based on bag-of-words assumption. The bag-of-words (or bag-of-phrases) representation cannot capture sequences and patterns that are very important to identify hate speech and offensive contents in tweets. For example, if a tweet ends saying \emph{if you know what I mean}, there is a high chance that it is an offensive tweet, even though individual words are innocent. 

A LSTM model is popularly used in natural language processing research because of its effectiveness of handling sequences in text datasets. Empirical results in Table \ref{tab:comp_models_val} show that it performed as a second best model. However, the sequence in a tweet can be highly impacted by the noise \cite{bashar2018misogynistic,xiang2012detecting}, consequently LSTM finds it difficult to identify the class. On the other hand, CNN can identify many small and large patterns in a tweet, if some of them are impacted by noise it can still use other patterns to identify the class.

\vspace{-4mm}
\section{Conclusion}
\vspace{-4mm}
We introduce an effective method for the task of hate speech and offensive content identification in Hindi. We propose a custom CNN architecture built on word vectors pre-trained on a relevant corpus from the task-specific domain. The proposed model was the top-ranked model in this task under the track. We conducted a series of experiments conducted using state-of-the-art models. Experimental results show that the contexts of hate speech and offensive content can be captured through transfer learning of word embeddings (a.k.a. word vectors) and those contexts can significantly improve the performance of hate speech and offensive content identification. We also observed that when transfer learning through word vectors is utilised, CNN performs better than LSTM because of the noisy nature of tweets. CNN can identify many small and large patterns in a tweet, if some of them gets altered by noise it can still use other patterns to identify the class of the tweet. On the other hand, LSTM uses the sequence of a tweet to identify its class, but noise in the tweet can alter the sequence and make it hard for LSTM to identify the class. 

%
%
%
\scriptsize
\bibliographystyle{splncs04}
\bibliography{References}

\end{document}